\documentclass[11pt,a4paper]{article}
\usepackage[hyperref]{emnlp2020}
\usepackage{times}
\usepackage{amsmath}
\usepackage{latexsym}
\usepackage{graphicx}
\usepackage{multirow}
\usepackage{amssymb}
\usepackage{booktabs}

\usepackage{microtype}

\aclfinalcopy 

\title{IGSQL: Database Schema Interaction Graph Based Neural Model for Context-Dependent Text-to-SQL Generation}

\author{Yitao Cai \and Xiaojun Wan \\
	Wangxuan Institute of Computer Technology, Peking University \\
	Center for Data Science, Peking University \\
	The MOE Key Laboratory of Computational Linguistics, Peking University \\
	{\tt \{caiyitao,wanxiaojun\}@pku.edu.cn} \\}

\date{}

\begin{document}
\maketitle
\begin{abstract}
Context-dependent text-to-SQL task has drawn much attention in recent years. Previous models on context-dependent text-to-SQL task only concentrate on utilizing historical user inputs. In this work, in addition to using encoders to capture historical information of user inputs, we propose a database schema interaction graph encoder to utilize historicalal information of database schema items. In decoding phase, we introduce a gate mechanism to weigh the importance of different vocabularies and then make the prediction of SQL tokens. We evaluate our model on the benchmark SParC and CoSQL datasets, which are two large complex context-dependent cross-domain text-to-SQL datasets. Our model outperforms previous state-of-the-art model by a large margin and achieves new state-of-the-art results on the two datasets. The comparison and ablation results demonstrate the efficacy of our model and the usefulness of the database schema interaction graph encoder. 
\end{abstract}

\section{Introduction}

The Text-to-SQL task aims to translate natural language texts into SQL queries. Users who do not understand SQL grammars can benefit from this task and acquire information from databases by just inputting natural language texts. Previous works \cite{li2014constructing,DBLP:journals/corr/abs-1711-04436,yu2018syntaxsqlnet,bogin2019global,huo2019graph} focus on context-independent text-to-SQL generation. However, in practice, users usually interact with systems for several turns to acquire information, which extends the text-to-SQL task to the context-dependent text-to-SQL task in a conversational scenario. Throughout the interaction, user inputs may omit some information that appeared before. This phenomenon brings difficulty for context-dependent text-to-SQL task. 

Recently, context-dependent text-to-SQL task has attracted more attention. \citet{suhr2018learning} conduct experiments on ATIS dataset \cite{dahl1994expanding}. Besides, two cross-domain context-dependent datasets SParC \cite{yu2019sparc} and CoSQL \cite{yu2019cosql} are released. Cross-domain means databases in test set differ from that in training set, which is more challenging. 

EditSQL \cite{zhang-emnlp19} is the previous state-of-the-art model on SParC and CoSQL datasets and it focuses on taking advantages of previous utterance texts and previously predicted query to predict the query for current turn. Table \ref{tab:example_editsql_error} shows the user inputs, ground truth queries and predicted queries of EditSQL for an interaction. In the second turn, EditSQL views ``Kacey" as the name of a dog owner. However, since the context of the interaction is about dogs, ``Kacey" should be the name of a dog. This example shows that a model using only historical information of user inputs may fail to keep context consistency and maintain thematic relations. 

According to \cite{yu2019sparc} and \cite{yu2019cosql}, to maintain thematic relations, users may change constraints, ask for different attributes for the same topic when they ask the next questions. Thus, database schema items (i.e., \textit{table.column}) in current turn should have relation with items in previous turn. For example, in Table \ref{tab:example_editsql_error}, the second question $x^2$ adds a constraint of the name and asks for the age of a dog instead of the numbers of all dogs. The corresponding database schema items \textit{Dogs.age} and \textit{Dogs.name} in $y^2$  belong to the same table as \textit{Dogs.*} in previous query $y^1$. Therefore, we propose to take historical information about database schema items into consideration.


In particular, we first construct a graph based on corresponding database, where graph nodes are database schema items and graph edges are primary-foreign keys and column affiliation. Short distance between graph nodes appearing in previous query and current query can reveal the context consistency since there is usually an edge between the different attributes of the same topic. We then propose a database schema interaction graph encoder to model database schema items together with historical items. Empirical results on two large cross-domain context-dependent text-to-SQL datasets - SParC and CoSQL show that our schema interaction graph encoder contributes to modeling context consistency and our proposed model with database schema interaction graph encoder substantially outperforms the state-of-the-art model.

\begin{table}[tbp]
    \small
    \centering
    \begin{tabular}{cl}
        \toprule
        $x^1$& how many dogs on the table \\
        \midrule
        $\Tilde{y}^1$ & SELECT  count ( * )  FROM  Dogs \\
        $y^1$& SELECT  count ( * )   FROM  Dogs \\
        \midrule
        $x^2$ & what is the age of Kacey \\
        \midrule
        $\Tilde{y}^2$ & SELECT  T2.age  FROM  owners  as  T1  JOIN  Dogs \\
        & AS  T2 ON T1.owner\_id = T2.owner\_id WHERE \\
        & T1.first\_name = 1 \\
        $y^2$ & SELECT age FROM dogs  WHERE  name = \\
        & ``Kacey" \\
        \midrule
        $x^3$ & which dog is highest weight on table \\
        & -- Do you want the name of the dog with the \\
        & highest weight? \\
        & -- exactly \\
        \midrule
        $\Tilde{y}^3$ & SELECT name FROM dogs ORDER BY weight \\ 
        & DESC limit 1 \\
        $y^3$ & SELECT name FROM dogs ORDER BY weight \\
        & DESC limit 1 \\
        \midrule
        $x^4$ & What is the size code of BUL \\
        & -- Did you mean the size code of dogs with a breed\\
        & \quad code BUL? \\
        & -- exactly \\
        \midrule
        $\Tilde{y}^4$ & SELECT size\_code FROM dogs WHERE \\
        & breed\_code = 1 \\
        $y^4$ & SELECT size\_code FROM dogs WHERE \\
        & breed\_code = ``BUL" \\
        \bottomrule
    \end{tabular}
    \caption{An example interaction. $x^i$ is the input sequence in $i$-th turn and $y^i$ is the corresponding ground truth query. $\Tilde{y}^i$ means that query is predicted by a model, which is EditSQL here.}
    \label{tab:example_editsql_error}
\end{table}

Our main contributions are summarized as follows: 

\begin{itemize}
    \item Previous models failed to keep context consistency and predict queries in a conversation scenario. To remedy this, we propose a database schema interaction graph encoder for database schema encoding and it can keep context consistency for the context-dependent text-to-SQL task. Our implementations are public available \footnote{https://github.com/headacheboy/IGSQL}.
    \item Our model with the database schema interaction graph encoder achieves new state-of-the-art performances on development and test sets of two cross-domain context-dependent text-to-SQL datasets, SparC and CoSQL.
\end{itemize}

\section{Related Work}

Many studies have focused on context-independent text-to-SQL task. \citet{DBLP:journals/corr/abs-1709-00103} split the vocabulary and use reinforcement learning. \citet{DBLP:journals/corr/abs-1711-04436} propose a sketched-based model, which decomposes the token prediction process into SELECT-clause prediction and WHERE-clause prediction, aiming at taking previous predictions into consideration. \citet{yu2018syntaxsqlnet} further employ a tree-based SQL decoder so as to decode SQL queries with the help of SQL grammar. In order to encode database schemas, schemas are regarded as graphs and graph neural networks have been applied \cite{bogin-etal-2019-representing,bogin2019global}. \citet{guo2019towards} design an intermediate representation to bridge the gap between natural language texts and SQL queries. \citet{choi2020ryansql} utilize a sketch-based slot filling approach to synthesize SQL queries. \citet{wang2019ratsql} attempt to align the database columns and their mentions in user inputs by using a relation-aware self attention. 

Recently, context-dependent text-to-SQL task has drawn people's attention. In-domain context-dependent benchmarks ATIS \cite{suhr2018learning} have been proposed. For ATIS, \citet{suhr2018learning} utilize a sequence to sequence framework. Besides, they introduce an interaction-level encoder for incorporating historical user inputs and a segment copy mechanism to reduce the length of generation. Later, two large and complex cross-domain context-dependent dataset SParC \cite{yu2019sparc} and CoSQL \cite{yu2019cosql} are proposed. In order to tackle cross-domain context-dependent text-to-SQL task, \citet{zhang-emnlp19} propose the EditSQL model in order to capture features from historical user inputs, variant database schemas and previously predicted SQL query. \citet{liu2020far} further evaluate context modeling methods and apply a grammar-based decoder. EditSQL achieves the state-of-the-art performance on the two cross-domain datasets.  Compared to EditSQL, our work further explore a new way to employ historical information of database schemas.

\begin{figure*}[ht]
\centering
\includegraphics[scale=0.5]{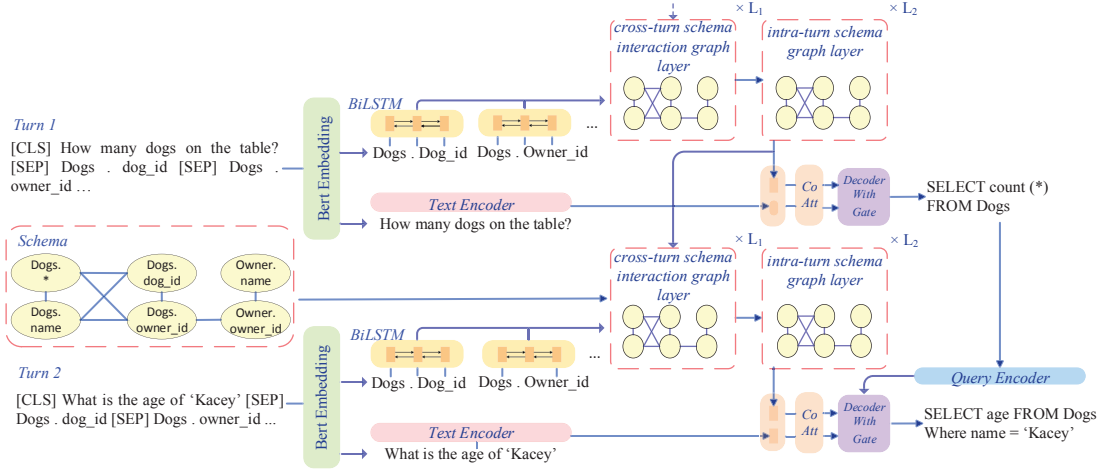}
\caption{Overview of our IGSQL model. Modules with the same color share the same parameters.}
\label{overview}
\end{figure*}

\section{Problem Setup}

We define $X$ as a series of natural language utterances of an interaction (i.e., user inputs), $Y$ as corresponding ground-truth SQL queries, $S$ as the set of database schema items (\textit{table.column}) and $R$ as the set of relations between schema items (primary-foreign keys and column affiliation). Let $X = \{x^1, x^2, ..., x^{|X|}\}$, where $|X|$ is the number of utterances. $x^i$ is the $i$-th utterance and $x^i_j$ is the $j$-th token of it. $y^i$ is the $i$-th SQL query corresponding to $x^i$ and $y^i_j$ is the $j$-th token of $y^i$. $S$ consists of schema items $\{S^1,...,S^{|S|}\}$, where $|S|$ is the number of database schema items. At turn $i$, the model should make use of current and previous utterances $\{x^1,x^2,...,x^i\}$, database schema items $S$ and their relations $R$ to predict a SQL query $\Tilde{y}^i$. The objective of the model is to maximize the probability of $\prod_{i=1}^{|I|}P(y^i|x^1,x^2,...,x^i)$.

\section{IGSQL Model}

Our model adopts an encoder-decoder framework with attention mechanism. Figure \ref{overview} shows the architecture of our model. The model have four main components: (1) a database schema interaction graph encoder, which consists of cross-turn schema interaction graph layers and intra-turn schema graph layers, (2) a text encoder that captures historical information of user inputs, (3) a co-attention module that updates outputs of text encoder and database schema interaction graph encoder, and (4) a decoder with a gated mechanism to weight the importance of different vocabularies. In addition, the model also uses BERT embedding. 

We will first introduce the BERT embedding in Section \ref{sec:bert}, and then introduce our database schema interaction graph encoder in Section \ref{model:graph}, text encoder and co-attention module in Section \ref{model:enc_coatt} and decoder in Section \ref{model:dec}. 

\subsection{BERT Embedding}
\label{sec:bert}

BERT \cite{devlin2019bert} is a pre-trained language model. Employing BERT output as embeddings of user inputs and database schema items has proved effective in context-dependent text-to-SQL task \cite{Hwang2019WikiSQL,guo2019towards,wang2019ratsql,choi2020ryansql}. Therefore, we leverage BERT to get the embeddings of user inputs and database schema items as other context-dependent text-to-SQL models do. We concatenate user inputs and database schema items by separating with a ``[SEP]" token following \cite{Hwang2019WikiSQL}. The output of BERT model is used as the embeddings of user inputs and schema items. 

\subsection{Database Schema Interaction Graph Encoder}
\label{model:graph}

As shown in Table \ref{tab:example_editsql_error}, previous model mistakes ``Kacey" as the name of a dog owner. However, the interaction is all about dogs and ``Kacey" should be the name of a dog. It shows that previous model does not perform well in modeling context consistency of an interaction. 

For two database schema items appearing in two adjacent turns, short distance of items in the graph can reveal the context consistency. For example, the distance between \textit{Dogs.*} \footnote{\textit{table.*} is considered a special column in table.} and correct item \textit{Dogs.name} is 1. Distance between \textit{Dogs.*} and wrong item \textit{owners.name} is 3. 

Therefore, we propose a database schema interaction graph encoder based on the database schema graph, attempting to model context consistency by using historical schema representations. The database schema interaction graph encoder consists of $L_{1}$ cross-turn schema interaction graph layers and $L_{2}$ intra-turn schema graph layers ($L_{1}$ and $L_{2}$ are hyper-parameters). Cross-turn schema interaction graph layers update schema item representations by using that in previous turn. Intra-turn schema graph layers further aggregate adjacent item representations in the same turn.

\subsubsection{Graph Construction and Schema Items Encoding}
\label{sec:graph_construction}

We first introduce how we construct a graph based on database schema. We use database schema items as nodes. Each node has an edge linking to itself. There is an undirected edge between node $t$ and node $j$ according to relation set $R$ if one of the following condition is satisfied: 1) node $t$ and node $j$ are the foreign-primary key pair; 2) node $t$ and node $j$ belong to the same table. We define the edge set as $E$.

A schema item \textit{table.column} is divided into \textit{``table"}, \textit{``."} and \textit{``column"}. We use a BiLSTM with BERT embedding to encode tokens and average hidden state vectors of BiLSTM as the embedding of the schema item. The embedding of the $j$-th schema item at $i$-th turn is noted as $r^i_j$.

\begin{figure}[tbp]
\centering
\includegraphics[scale=0.7]{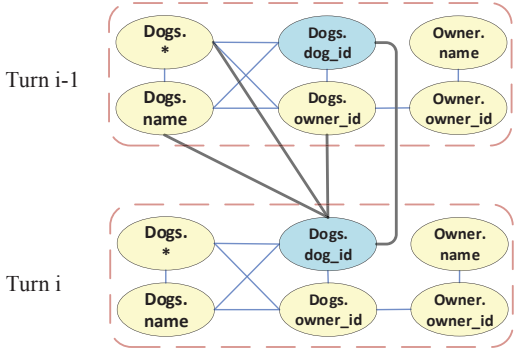}
\caption{Database Schema Interaction Graph. We add black edges into the graph when we want to update the representation of the blue node (\textit{Dogs.dog\_id}) at turn $i$.}
\label{fig:IG}
\end{figure}

\subsubsection{Cross-turn Schema Interaction Graph Layer} Figure \ref{fig:IG} shows an example of the database schema interaction graph. The graph only allows node $t$ in previous turn to update node $j$ in current turn, when the distance between node $t$ and node $j$ in the original graph constructed in Section \ref{sec:graph_construction} is less than or equal to 1. For example, if we want to update the representation of \textit{Dogs.dog\_id} at turn $i$, we add edges linking \textit{Dogs.*}, \textit{Dogs.name}, \textit{Dogs.owner\_id} and \textit{Dogs.dog\_id} at turn $i-1$ to \textit{Dogs.dog\_id} at turn $i$.

Note that we have $L_{1}$ cross-turn schema interaction graph layers for turn $i$. At the \textit{l}-th layer, we obtain updated representation $z^{i,l}_t$ of the $t$-th schema item by using attention on outputs of the $L_{2}$ intra-turn schema graph layers at previous turn $\{g^{i-1,L_2}_t\}_{t=1}^{|S|}$ (which will be introduced in next subsection) and representations of previous layer $\{z^{i,l-1}_t\}_{t=1}^{|S|}$. We use item embedding $r^i_t$ as the initial representation $z^{i,0}_t$. For simplicity, we omit turn index $i$ and layer index \textit{l} in the formulas of attention mechanism except the input $z^{i,l-1}_t$, $g^{i-1,L_2}_t$ and output $z^{i,l}_t$.

At the $l$-th layer, we first use a feed-forward neural network with leakyReLU activation function for non-linear transformation. We use FFN to denote the feed-forward neural network with leakyReLU activation function.

\begin{equation}
    \begin{split}
        &u_t={\rm FFN}(z^{i,l}_t) \\
        &\hat{u}_t = {\rm FFN}(g^{i-1,L_2}_t)
    \end{split}
\end{equation}

We then apply attention mechanism as follows. 

\begin{equation}
\label{equ:graph_att}
    \begin{split}
        &\xi_{t,j} =\left\{ 
        \begin{aligned}\
        &(u_t)^TW_1u_j / \sqrt{d_1}, &[t,j] \in E \\
        &-\infty, &[t,j] \notin E
        \end{aligned}
        \right.\\
        &\hat{\xi}_{t,j} =\left\{ 
        \begin{aligned}\
        &(u_t)^TW_2\hat{u}_j / \sqrt{d_1}, &[t,j] \in E \\
        &-\infty, &[t,j] \notin E
        \end{aligned}
        \right.\\
        &\alpha_{t,j}=\frac{exp(\xi_{t,j})}{\sum_{v}exp(\xi_{t,v})+\sum_kexp(\hat{\xi}_{t,k})} \\
        &\hat{\alpha}_{t,j}=\frac{exp(\hat{\xi}_{t,j})}{\sum_vexp(\xi_{t,v})+\sum_kexp(\hat{\xi}_{t,k})} \\
        &\Tilde{u}_t = \sum_{j}\alpha_{t,j}u_j + \sum_{j}\hat{\alpha}_{t,j}\hat{u}_j \\
    \end{split}
\end{equation}

\noindent where $d_1$ is the dimension of $u_t$. $W_1$ and $W_2$ are weight matrices. $\alpha_{t,j}$ and $\hat{\alpha}_{t,j}$ are the attention scores. $\Tilde{u}_t$ is the $t$-th output vector of attention.

Following \cite{vaswani2017attention,velivckovic2017graph}, we extend attention mechanism to multi-head attention. We also add a sub-layer of feed-forward neural network with residual connection as in Transformer. 

\begin{equation}
    \begin{split}
        & z^{i,l}_t = z^{i,l-1}_t + {\rm FFN}( z^{i,l-1}_t+\Tilde{u}_t)
    \end{split}
\end{equation}

\noindent where $z^{i,l}_t$ is the final output of layer \textit{l}. There are $L_1$ cross-turn schema interaction graph layers  and thus $z^{i,L_1}_t$ is the final output of cross-turn schema interaction graph layers for the $t$-th schema item. 

\subsubsection{Intra-turn Schema Graph Layer} There are $L_2$ intra-turn schema graph layers following cross-turn schema interaction graph layers. In each intra-turn schema graph layer, we use almost the same attention mechanism as in the cross-turn schema interaction graph layer, except that we use the original graph constructed in Section  \ref{sec:graph_construction}. Since the original graph does not contain nodes in previous turn, the intra-turn schema graph layer can only update node representation by aggregating adjacent node representations in the same turn. 

At each intra-turn schema graph layer layer \textit{l} of turn $i$, it takes output vectors in previous layer $g^{i,l-1}_t$ as inputs and its output is $g^{i,l}_t$. $g^{i,0}_t$ is $z^{i,L_1}_t$. We then use attention mechanism to aggregate information. We also add a sub-layer of FFN and residual connection. For simplicity, we omit the turn index $i$ and layer index \textit{l} in attention except input $g^{i,l-1}_t$ and output $g^{i,l}_t$.

\begin{equation}
    \begin{split}
        &\mu_t={\rm FFN}(g^{i,l-1}_t) \\
        &\tau_{t,j} =\left\{ 
            \begin{aligned}\
            &(\mu_t)^TW_3\mu_j / \sqrt{d_2}, &[t,j] \in E \\
            &-\infty, &[t,j] \notin E
            \end{aligned}
            \right . \\
        &\beta_{t,j}= \frac{exp(\tau_{t,j})}{\sum_k exp(\tau_{t,k})} \\
        &\Tilde{\mu}_t=\sum_{j}\beta_{t,j}\mu_j \\
        &g^{i,l}_t=g^{i,l-1}_t+{\rm FFN}(g^{i,l-1}_t+\Tilde{\mu}_t) 
    \end{split}
\end{equation}

\noindent where $W_3$ is a weight matrix and $d_2$ is the dimension of $\mu_t$. $\beta_{t,j}$ is the attention score of the $j$-th node to the $t$-th node. $\Tilde{\mu}_t$ is the attention output. $g^{i,l}_t$ is the output of $t$-th schema item at layer $l$ of turn $i$. Besides, We also extend attention to multi-head attention.

The final output of intra-turn schema graph layers for the $t$-th schema item is $g^{i,L_2}_t$.

\subsection{Text Encoder and Co-Attention Module}
\label{model:enc_coatt}

We use a BiLSTM to encode tokens of an utterance text with BERT embedding. In order to capture interaction history, we add an LSTM as interaction encoder and utilize turn-level attention, following \cite{zhang-emnlp19}. The final representation of the $t$-th token in utterance $i$ is denoted as $h^i_t$.

We also add a co-attention module between text tokens and schema items following \cite{zhang-emnlp19}. The schema item vector $\Tilde{g}^i_t$ used in decoding phase is the concatenation of $g^{i,L_2}_t$ and its corresponding attention vector over text. The representation of input text tokens $\Tilde{h}^i_t$ used in decoding phase is the concatenation of $h^i_t$ and its corresponding attention vector over schema items. Due to page limit, we omit the details here, which can be found in \cite{zhang-emnlp19}.

\subsection{Decoder}
\label{model:dec}

In decoding phase, we first encode previously predicted query with a BiLSTM. We then exploit a LSTM decoder with attention \cite{bahdanau2015neural}  to capture features from input text's token vectors, schema item vectors and previously predicted SQL query vectors. At $j$-th time step, We use attention on text token's vector $\Tilde{h}^i_t$, database schema vector $\Tilde{g}^i_t$ and previously predicted SQL token's vector $q_t$. We thus get three context vectors. The final context vector $c_j$ is the concatenation of these three context vectors.  

We follow \cite{suhr2018learning} to make prediction of SQL tokens based on SQL reserved words, database schema items and previous predicted SQL  tokens. We also add a gate mechanism to introduce the importance of these three vocabularies. For simplicity, we omit turn index $i$ in decoder step except $\Tilde{y}^i_j$. 

The gate mechanism is introduced to measure the importance of three vocabularies. 

\begin{equation}
\label{equ:intermediate}
    \begin{split}
        &\Tilde{o}_j = {\rm tanh}(W_o([o_j;c_j]+b_o)\\
        &\zeta_m = \sigma(W_m\Tilde{o}_j+b_m) \\
        &m \in \{res, sch, que\} 
    \end{split}
\end{equation}

\noindent where $o_j$ is the $j$-th hidden vector of the LSTM decoder. $c_j$ is the context vector. $[;]$ is the concatenation operator and $\Tilde{o}_j$ is the non-linear transformation of $[o_j;c_j]$. $\sigma$ is the sigmoid function. $res,sch,que$ represent SQL reserved words, database schema items and previously predicted SQL tokens respectively and $\zeta_{res},\zeta_{sch},\zeta_{que}$ represent the importance of these three kinds of tokens.

We then predict SQL tokens as follows. 

\begin{equation}
    \begin{split}
        &p_1(\Tilde{y}^i_j=w) = \frac{1}{Z}exp(\zeta_{res}\cdot \textbf{w}^T(W_{res}\Tilde{o}_j+b_{res})) \\
        &p_2(\Tilde{y}^i_j=S_t)= \frac{1}{Z} exp(\zeta_{sch}\cdot(\Tilde{g}^i_t W_{sch}\Tilde{o}_j) \\
        &p_3(\Tilde{y}^i_j=\Tilde{y}^{i-1}_t) = \frac{1}{Z} exp(\zeta_{que}\cdot (q_t W_{que}\Tilde{o}_j )) 
    \end{split}
\end{equation}

\noindent where \textbf{w} is the one-hot vector of word $w$. $q_t$ and $\Tilde{g}^{i}_t$ are query vector and schema item vector that are mentioned before. The final generation probability $p(\Tilde{y}^i_j)$ is $p_1(\Tilde{y}^i_j)+p_2(\Tilde{y}^i_j)+p_3(\Tilde{y}^i_j)$. $Z$ is the normalization factor that ensures $\sum_{v \in V} p(v)$ is 1, where $V$ is the whole vocabulary. The loss function is $\sum_i \sum_j -log(p(y^i_j))$

\begin{table}[tbp]
    \centering
    \begin{tabular}{ccc}
        \toprule
        &SParC & CoSQL\\
        \midrule
        cross-domain & \checkmark & \checkmark\\
        Interaction & 4298 & 3007 \\
        Train & 3034 & 2164\\
        Dev & 422 & 292\\
        Test & 842 & 551\\
        User Questions & 12726 & 15598 \\
        Databases & 200 & 200 \\
        Tables & 1020 & 1020 \\
        Vocab & 3794 & 9585  \\
        Avg Turn & 3.0 & 5.2 \\
        \bottomrule
    \end{tabular}
    \caption{Statistics of SParC, CoSQL}
    \label{tab:statistics}
\end{table}

\begin{table*}[htb]
    \small
    \centering
    \begin{tabular}{ccccccccc}
       \toprule
       \multirow{2}{*}{Method}
         &  \multicolumn{2}{c}{SParC Dev} & \multicolumn{2}{c}{SParC Test} & \multicolumn{2}{c}{CoSQL Dev} & \multicolumn{2}{c}{CoSQL Test}\\
         \cmidrule(r){2-3} \cmidrule(r){4-5} \cmidrule(r){6-7} \cmidrule(r){8-9}
         & Ques & Int & Ques & Int & Ques & Int & Ques & Int \\
       \midrule
       CD S2S & 21.9& 8.1& 23.2& 7.5& 13.8& 2.1& 13.9& 2.6\\
       SyntaxSQL-con & 18.5& 4.3& 20.2& 5.2& 15.1& 2.7& 14.1& 2.2\\
       EditSQL*  & 47.2 & 29.5 & 47.9  &25.3  &39.9 &12.3 & 40.8 & 13.7 \\
       IGSQL* & \textbf{50.7} & \textbf{32.5} & \textbf{51.2} & \textbf{29.5} & \textbf{44.1} & \textbf{15.8} & \textbf{42.5} & \textbf{15.0} \\
       \bottomrule
    \end{tabular}
    \caption{Results of models in SParC and CoSQL datasets. Ques means question match accuracy. Int means interaction match accuracy. * means that results are enhanced by BERT embedding.}
    \label{tab:result}
\end{table*}

\begin{table}[tb]
    \small
    \centering
    \begin{tabular}{ccccc}
       \toprule
       \multirow{2}{*}{Turns}
         &  \multicolumn{2}{c}{SParC} & \multicolumn{2}{c}{CoSQL} \\
         \cmidrule(r){2-3} \cmidrule(r){4-5}
         & EditSQL & IGSQL & EditSQL & IGSQL \\
       \midrule
       1 & 62.2 & 63.2 & 50.0 & 53.1 \\
       2 & 45.1 & 50.8 & 36.7 & 42.6 \\
       3 & 36.1 & 39.0 & 34.8 & 39.3 \\
       4 & 19.3 & 26.1 & 43.0 & 43.0 \\
       $\textgreater$4 & 0 & 0 & 23.9 & 31.0  \\
       \bottomrule
    \end{tabular}
    \caption{Exact match accuracy w.r.t. turn number on development sets.}
    \label{tab:turn_acc}
\end{table}

\begin{table}[tb]
    \small
    \centering
    \begin{tabular}{ccccc}
       \toprule
       \multirow{2}{*}{Hardness}
         &  \multicolumn{2}{c}{SParC} & \multicolumn{2}{c}{CoSQL} \\
         \cmidrule(r){2-3} \cmidrule(r){4-5}
         & EditSQL & IGSQL & EditSQL & IGSQL \\
       \midrule
       Easy & 68.8 & 70.9 & 62.7 & 66.3 \\
       Medium & 40.6 & 45.4 & 29.4 & 35.6 \\
       Hard & 26.9 & 29.0 & 22.8 & 26.4 \\
       Extra & 12.8 & 18.8 & 9.3 & 10.3 \\
       \bottomrule
    \end{tabular}
    \caption{Exact match accuracy w.r.t. different hardness level on development sets.}
    \label{tab:hard_acc}
\end{table}

\begin{table*}[htb]
    \small
    \centering
    \begin{tabular}{ccccc}
       \toprule
       \multirow{2}{*}{Method}
         &  \multicolumn{2}{c}{SParC} & \multicolumn{2}{c}{CoSQL} \\
         \cmidrule(r){2-3} \cmidrule(r){4-5}
         & Ques Match & Int Match & Ques Match & Int Match \\
       \midrule
       IGSQL & 50.7 & 32.5 & 44.1 & 15.8 \\
       w/o cross-turn schema interaction graph layer  & 47.6(-3.1) & 29.5(-3.0) & 41.9(-2.2) & 14.0(-1.8) \\
       w/o intra-turn schema graph layer & 50.2(-0.5) & 31.1(-1.4) & 42.9(-1.2) & 14.0(-1.8) \\
       GRU interaction layer & 48.2(-2.5) & 29.2(-3.3) & 41.0(-3.1) & 14.1(-1.7) \\
       Fully-connected interaction layer & 48.2(-2.5) & 29.0(-3.5) & 42.0(-2.1) & 13.0(-2.8) \\
       \bottomrule
    \end{tabular}
    \caption{Ablation study on development sets. Numbers in brackets are performance differences compared to IGSQL. }
    \label{tab:ablation study}
\end{table*}

\section{Implementation Details}

We use Adam optimizer \cite{kingma2014adam} to optimize the loss function. The initial learning rate except BERT model is 1e-3, while the initial learning rate of BERT model is 1e-5. We use learning rate warmup over the first 1000 steps. The learning rate will be multiplied by 0.8 if the loss on development set increases and the token accuracy on development set decreases. The number of cross-turn schema interaction graph layer $L_1$ is 2, while the number of intra-turn schema graph layer $L_2$ is 1. The dimensions $d_1$ and $d_2$ are both 300. For encoder and decoder, the hidden size of the one layer LSTM and BiLSTM are 300. Besides, we use batch re-weighting to reweigh the loss function following \cite{suhr2018learning}. For BERT embedding, following EditSQL, we use the pre-trained BERT base model in order to make fair comparison. 

\section{Experiments}

\subsection{Experiment Setup}

\textbf{Datasets.}
We conduct experiments on two large-scale cross-domain context-dependent SQL generation datasets, SParC \cite{yu2019sparc} and CoSQL \cite{yu2019cosql}. In comparison with previous context-dependent dataset ATIS \cite{dahl1994expanding}, SParC and CoSQL are more complex since they contain more databases and adopt a cross-domain task setting, where the databases of training set differ from that of development set and test set. Statistics of SParC and CoSQL are shown in Table \ref{tab:statistics}.

\textbf{Evaluation Metrics.} \citet{yu2018spider} introduce exact set match accuracy to replace string match accuracy by taking queries with same constraints but different orders as the same query. In SParC and CoSQL, we use question match accuracy and interaction match accuracy as evaluation metrics. Question match accuracy is the average exact set match accuracy over all questions, while interaction match accuracy is the average exact set match accuracy over all interactions.

\begin{table*}[tbp]
    \small
    \centering
    \begin{tabular}{cl}
        \toprule
        $x^1$& Which cartoon aired first? \\
        \midrule
        EditSQL & SELECT title FROM cartoon ORDER BY original\_air\_date LIMIT 1 \\
        IGSQL & SELECT title FROM cartoon ORDER BY original\_air\_date LIMIT 1 \\
        $y^1$& SELECT title FROM cartoon ORDER BY original\_air\_date LIMIT 1 \\
        \midrule
        $x^2$ & What was the last cartoon to air? \\
        \midrule
        EditSQL & SELECT T1.title FROM cartoon \textcolor{red}{AS T1 JOIN tv\_channel AS T2 ON T1.channel = T2.id JOIN tv\_series} \\
        & \textcolor{red}{AS T3 ON T2.id = T3.channel ORDER BY T3.air\_date} LIMIT 1 \\
        IGSQL & SELECT title FROM cartoon \textcolor{red}{ORDER BY original\_air\_date DESC} LIMIT 1 \\
        $y^2$ & SELECT title FROM cartoon \textcolor{red}{ORDER BY original\_air\_date DESC} LIMIT 1 \\
        \midrule
        $x^3$ & What channel was it on? \\
        \midrule
        EditSQL & SELECT channel FROM \textcolor{red}{tv\_series ORDER BY air\_date} LIMIT 1 \\
        IGSQL & SELECT channel FROM \textcolor{red}{cartoon ORDER BY original\_air\_date DESC} LIMIT 1\\
        $y^3$ & SELECT channel FROM \textcolor{red}{cartoon ORDER BY original\_air\_date DESC} LIMIT 1\\
        \midrule
        $x^4$ & What is the production code? \\
        \midrule
        EditSQL & select T1.production\_code FROM cartoon \textcolor{red}{AS T1 JOIN tv\_channel AS T2 ON T1.channel = T2.id JOIN} \\
        & \textcolor{red}{tv\_series AS T3 ON T2.id = T3.channel ORDER BY T3.air\_date} LIMIT 1 \\
        IGSQL & SELECT production\_code FROM cartoon \textcolor{red}{ORDER BY original\_air\_date DESC} LIMIT 1 \\
        $y^4$ & SELECT production\_code FROM cartoon \textcolor{red}{ORDER BY original\_air\_date DESC} LIMIT 1 \\
        \bottomrule
    \end{tabular}
    \caption{An example of an interaction in CoSQL. $x^i$ is the input sequence at $i$-th turn and $y^i$ is the corresponding ground truth query. We show the predictions of EditSQL and IGSQL and mark the differences with red color.}
    \label{tab:case_study}
\end{table*}    

\noindent
\textbf{Baseline Models.} We compare our model with following baseline models.

\begin{itemize}
    \item \textbf{Context dependent Seq2Seq (CD S2S).} This model is originated in \cite{suhr2018learning} for ATIS dataset. \citet{yu2019sparc} adapt this model to cross-domain setting by adding a BiLSTM to encode schema items and modifying the decoder to generate different schema items according to databases.
    \item \textbf{SyntaxSQL-con.} This model is originated in \cite{yu2018syntaxsqlnet}, which utilizes SQL grammars for decoder. \citet{yu2019sparc} adapt this model to context-dependent setting by adding LSTM encoders to encode historical user inputs and historical SQL queries. 
    \item \textbf{EditSQL.} The model is proposed by \cite{zhang-emnlp19}. In addition to modules for encoding historical user inputs and corresponding SQL queries, it also contains a copy mechanism to copy tokens from previous SQL queries. 
\end{itemize}

\subsection{Experiment Results}

Results of these baseline models and our proposed IGSQL model are shown in Table \ref{tab:result}. Our model surpasses the previous state-of-the-art model EditSQL. IGSQL achieves substantial improvement on question match accuracy by 3.5, 3.3 points on SParC development and test sets and 4.2, 1.7 points on CoSQL development and test sets, respectively. As for interaction match accuracy, IGSQL improves by 3, 4.2 points on SParC development and test sets, and 3.5, 1.3 points on CoSQL development and test sets. Results demonstrate the effectiveness of our model. 

Table \ref{tab:turn_acc} shows the exact match accuracy of interaction with respect to different turn number. In both datasets, performances on interactions with one turn improve less. In SParC, performances on interactions with two turns and four turns improve the most, while in CoSQL, performances on interaction with two turns and larger than four turns improve the most. These results demonstrate that our database schema interaction graph encoder contributes to modeling schema items in conversational scenarios. 

Table \ref{tab:hard_acc} lists the exact match accuracy with respect to different hardness level. Results in the table show that performance at each hardness level improves. The results indicate that capturing historical database schema information can not only improve the accuracy of easy questions, but also answer harder questions more accurately.

\subsection{Ablation Study}

In order to verify the usefulness of our database schema interaction graph encoder, we conduct several ablation experiments as follows.

\noindent
\textbf{w/o cross-turn schema interaction graph layer.} In this experiment, we discard cross-turn schema interaction graph layers. In this setting, our model cannot encode historical database schema information. 

\noindent
\textbf{w/o intra-turn schema graph layer.} In this experiment, we discard intra-turn schema graph layers to examine whether these layers are useful. 

\noindent
\textbf{GRU interaction layer.} One of the most common way to employ historical information of database schema items is to update node representation directly from historical vector of the same node. For example, in Figure \ref{fig:IG}, we can use a GRU by taking representation of \textit{Dogs.dog\_id} at turn $i-1$ and its BERT embedding at turn $i$ as input. The output of GRU is the vector of \textit{Dogs.dog\_id} at turn $i$. In this experiment, we use a GRU to replace cross-turn schema interaction graph layers. 

\noindent
\textbf{Fully-connected interaction layer.} To examine the effectiveness of our design of schema interaction graph, we make experiment that replaces the schema interaction graph with fully connected graph. Taking Figure \ref{fig:IG} as an example, to update representation of blue node at turn $i$, there are edges connecting blue node at turn $i$ to all nodes at turn $i-1$. 

Since the test sets of SParC and CoSQL are not public, we carry out the ablation experiments only on development sets of these two datasets. Table \ref{tab:ablation study} shows the results of ablation experiments. Our full model achieves about 2 points improvement compared with the model without cross-turn schema interaction graph layers and the model with GRU interaction layer. Besides, our model achieves about 1 point improvement compared with the model without intra-turn schema graph layers. These results indicate that our cross-turn and intra-turn schema graph layers are very helpful.

The difference between cross-turn schema interaction graph layer and fully-connected interaction layer is how we add edges between nodes at turn $i-1$ and turn $i$. Compared to fully-connected interaction layer, the schema interaction graph introduces a distance restriction when adding edges. Our model with schema interaction graph performs substantially better, which shows that our design of schema interaction graph can significantly help our model to keep context consistency. 

\subsection{Case Study}

In Table \ref{tab:case_study}, we show an interaction with four turns. We also provide the predictions of EditSQL and IGSQL and mark the differences with red color. After the first turn, EditSQL confuses \textit{cartoon.original\_air\_date} with \textit{tv\_series.air\_date}. Our proposed IGSQL model successfully obtains answers in the correct order by taking historical information of database schema items into account.

\section{Conclusion and Future work}

In this paper, we focus on context-dependent cross-domain SQL generation task. We find that previous state-of-the-art model only takes historical user inputs and previously predicted query into consideration, but ignores the historical information of database schema items. Thus we propose a model named IGSQL to model database schema items in a conversational scenario. Empirical results demonstrate the efficacy of our model. We also conduct ablation experiments to reveal the significance of our database schema interaction graph encoder. For future work, we 
will explore methods attempting to solve hard and extra hard questions. 

\section*{Acknowledgments}

This work was supported by National Natural Science Foundation of China (61772036), Beijing Academy of Artificial Intelligence (BAAI) and Key Laboratory of Science, Technology and Standard in Press Industry (Key Laboratory of Intelligent Press Media Technology). We appreciate the anonymous reviewers for their helpful comments. Xiaojun Wan is the corresponding author.

\bibliography{emnlp2020}
\bibliographystyle{acl_natbib}

\end{document}